\documentclass[11pt]{article} 
\usepackage[preprint]{acl}  
\usepackage{algorithm}

\usepackage{times}
\usepackage{latexsym}
\usepackage{url}            
\usepackage{booktabs}       
\usepackage{amsfonts}       
\usepackage{nicefrac}       
\usepackage{microtype}      
\usepackage{latexsym}
\usepackage{amssymb}
\usepackage{amsmath}
\usepackage{booktabs}
\usepackage{enumerate}
\usepackage{graphicx}
\usepackage{subfigure}
\usepackage{xspace}
\usepackage{float}
\usepackage{bm}
\usepackage{multirow}
\usepackage{booktabs}
\usepackage{color}
\usepackage{framed}
\usepackage{iitem}
\usepackage{amssymb}
\usepackage{pifont}
\usepackage{arydshln}
\usepackage{enumitem}
\usepackage{algpseudocode}
\usepackage{array}
\usepackage{tabularx}
\usepackage{booktabs}
\usepackage{float}
\usepackage[T1]{fontenc}
\usepackage{colortbl}
\definecolor{maroon}{cmyk}{0,0.87,0.68,0.32}

\usepackage[utf8]{inputenc}

\usepackage{microtype}

\definecolor{bittersweet}{rgb}{1.0, 0.44, 0.37}
\definecolor{mygreen}{rgb}{0.29, 0.7, 0.48}
\usepackage{pifont}
\definecolor{demphcolor}{RGB}{144,144,144}

\definecolor{mygray}{gray}{0.4}
\definecolor{darksalmon}{rgb}{0.91, 0.59, 0.48}
\definecolor{emerald}{rgb}{0.31, 0.78, 0.47}
\definecolor{green(pigment)}{rgb}{0.0, 0.65, 0.31}
\definecolor{amaranth}{rgb}{0.9, 0.17, 0.31}
\definecolor{iris}{rgb}{0.35, 0.31, 0.81}
\definecolor{uu}{rgb}{0.95, 0.51, 0.51}
\definecolor{spirodiscoball}{rgb}{0.06, 0.75, 0.99}
\usepackage{svg}

\usepackage{cleveref}

\usepackage{asymptote}
\usepackage{amsmath}
\usepackage{xspace}

\definecolor{ada_blue}{rgb}{0,205,205}
\definecolor{glt_red}{rgb}{109,205,255}
\definecolor{MorandiBlue}{RGB}{118,134,146}

\definecolor{demphcolor}{RGB}{144,144,144}
\definecolor{mygray}{gray}{0.4}
\definecolor{autopurple}{HTML}{7030A0}
\definecolor{dyna_yellow}{HTML}{BF9000}
\definecolor{adaptive_blue}{HTML}{0070C0}
\definecolor{darkgrey}{RGB}{120,120,120}
\definecolor{mygrey}{RGB}{200,200,200}

\usepackage{makecell}
\usepackage{tabulary}
\usepackage{tabularx}
\usepackage[most]{tcolorbox}

\definecolor{myblue}{HTML}{00CDCD}
\definecolor{champagne}{rgb}{0.74, 0.83, 0.9}
\definecolor{champagne}{rgb}{0.97, 0.91, 0.81}

\usepackage{inconsolata}
\usepackage{multirow}
\newcommand{\method}{SAI-DPO\xspace}
\usepackage{newfloat}
\usepackage{listings}

\DeclareCaptionStyle{ruled}{labelfont=normalfont,labelsep=colon,strut=off} 
\floatstyle{ruled}
\newfloat{listing}{tb}{lst}{}
\floatname{listing}{Listing}
\title{Dynamic Sampling that Adapts: Self-Aware Iterative Data \\ Persistent Optimization for Mathematical Reasoning}
\author{
Jun Rao$^{1}$,
 Xuebo Liu$^{1}$\thanks{~Corresponding Authors.},
 Hexuan Deng$^{1}$,~
 Zepeng Lin$^{1}$,~\\
 \bf{Zixiong Yu}$^{2}$,~
 \bf{Jiansheng Wei}$^{2}$,~
  \bf{Xiaojun Meng}$^{2}$$^{*}$, and
    \bf{Min Zhang}$^{1}$\\
    \textsuperscript{\rm1}Institute of Computing and Intelligence, Harbin Institute of Technology, Shenzhen, China \\
    \textsuperscript{\rm2}Huawei Large Model Data Technology Lab~~~
    \\
    \texttt{\{rao7jun,zepenglin11,hxuandeng\}@gmail.com, \{liuxuebo,zhangmin2021\}@hit.edu.cn}\\
    \texttt{yuzx19@tsinghua.org.cn},~
    \texttt{\{weijiansheng,xiaojun.meng\}@huawei.com}
    }

\usepackage{bibentry}
\usepackage[switch]{lineno}

\begin{document}

\maketitle

\begin{abstract}
In mathematical reasoning, data selection strategies predominantly rely on static, externally defined metrics, which fail to adapt to the evolving capabilities of models during training. This misalignment limits the efficiency of Supervised Fine-Tuning and Reinforcement Learning. To bridge this gap, we introduce SAI-DPO (Self-Aware Iterative Data Persistent Optimization), a dynamic sampling framework that aligns training data with the model's intrinsic competence. SAI-DPO operationalizes two novel metrics: Knowledge Semantic Alignment for targeting domain weaknesses, and Self-Aware Difficulty, derived from pass rates and reasoning path characteristics, to gauge instance complexity relative to the model's current state. 
By iteratively recalibrating the data distribution based on real-time feedback, SAI-DPO dynamically aligns training samples with the model's evolving competence, ensuring the data remains strictly relevant to the model's current capability level.
Extensive experiments on eight benchmarks (including AIME24 and AMC23) demonstrate that SAI-DPO outperforms static baselines at most nearly 6 points, achieving state-of-the-art efficiency with significantly less data.  
\end{abstract}


\section{Introduction}

\begin{figure}[t]
    \centering
    \includegraphics[width=1.0\linewidth]{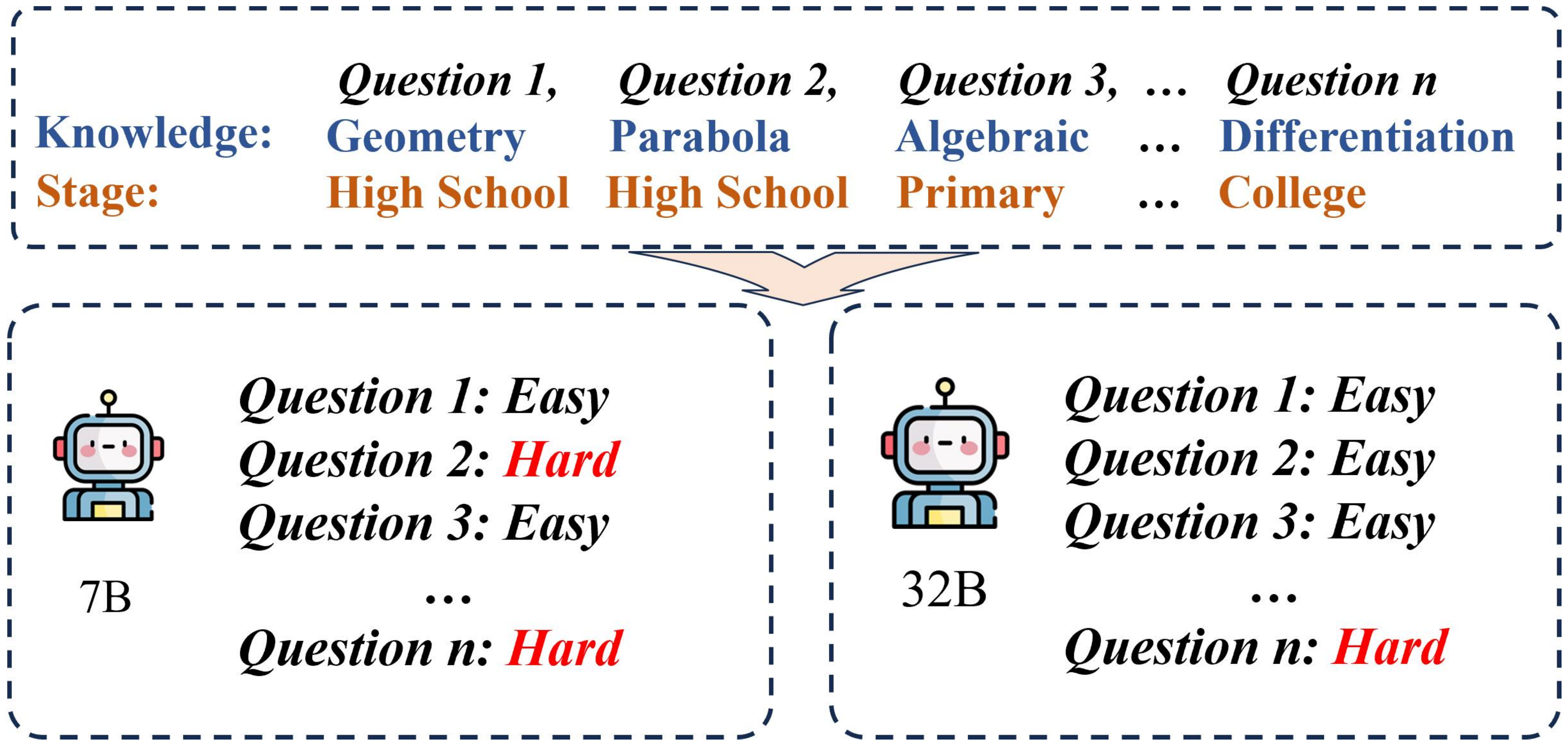}
    \caption{The difficulty levels and knowledge scopes of problems solvable by different models vary significantly. Basic knowledge points are manageable for small models, while complex or advanced content poses challenges. Even for a large model, such advanced problems (Differentiation) remain difficult, highlighting the importance of aligning difficulty with capabilities. Therefore, the difficulty definition should rely on the current capabilities of the model.
    }
    \label{intro}
\end{figure}
Recent advances in Large Language Models (LLMs), particularly in reasoning tasks~\cite{openai2024openaio1card,guo2025deepseek}, highlight the critical role of high-quality data. 
However, current data selection paradigms remain largely static, relying on fixed datasets or external difficulty scorers. 
This creates a fundamental disconnection: as a model learns, what was once ``hard'' becomes ``easy'', rendering static datasets progressively inefficient. 
Continued training on trivial samples yields diminishing returns, while overly complex samples may induce hallucinations.

Current works mainly focus on Supervised Fine-Tuning (SFT)~\cite{alpaca,commonIT,zhou2023lima,liu2024selectit,zeng-etal-2025-bridging} after data filtering~\cite{muennighoff2025s1simpletesttimescaling,ye2025limoreasoning} or online reinforcement learning algorithms~\cite{grpo,ppo,deng2026rearl}. 
Most of these methods are static, failing to adaptively select suitable data for continuous training based on the model's current capabilities, thereby limiting the sustainable improvement of its reasoning abilities.

As shown in Figure \ref{intro}, different models have varying capabilities, thus leading to differences in their discrimination of the questions.
Although some existing works~\cite{team2025kimi,zhou2026look} have addressed the impact of difficulty on models~\cite{ke2025aquiltweavinglogicselfinspection,liu2024selectit}, the related metrics remain unclear.
To address the issue of the lack of dynamic adaptive training for reasoning data, we propose the \method (Self-Aware Iterative Data Persistent Optimization) algorithm for mathematical reasoning.
This algorithm dynamically selects training data that matches the model's current competence (\textit{Self-Aware Difficulty}) and weaknesses (\textit{Knowledge Semantic Alignment}), enhancing its reasoning abilities through iterations. Using the defined metrics, the algorithm dynamically selects data and filters low-quality inputs to enhance training efficiency.

We conducted extensive experiments to explore the defined metric, data acquisition strategy, and the gradual improvement through iterative training. 
The experiments were carried out on 8 existing public mathematical test sets and 4 public models (Qwen2.5-7B-Math-Base, Qwen2.5-Math-7B-SFT, Llama3.1-8B-Instruct and Qwen3-8B). 
Our approach not only achieves better performance compared to the original DPO but also accelerates the training process.
And compared to some current common strategies, such as externally defined difficulty~\cite{ye2025limoreasoning} and curriculum learning~\cite{chen2025self,hong2025glm}, our strategy has better results.
Our results show that externally defined difficulty does not align with what is difficult for the model, and it is better to train with the model's defined difficulty.
Our main contributions are as follows:
\begin{itemize}

\item We propose a Dynamic Data Acquisition strategy that clusters knowledge tags to systematically target specific weakness domains.

\item We formulate a Self-Aware Difficulty Metric that integrates statistical priors (pass rate) with cognitive load indicators (step count and length), providing a nuanced view of model competence.

\item We demonstrate through extensive experimentation that aligning data difficulty with model capability yields superior performance, improving accuracy on competition-level benchmarks (AIME24 and AMC23) by nearly 4 points over strong baselines.
\end{itemize}

\section{Related Work}
\subsection{Post-training Preference Optimization}
In the post-training stage, many RL algorithms improve model performance by aligning the model's output objectives with human preferences-specifically, by increasing the probability of generating high-quality responses and decreasing the probability of producing low-quality ones.
A common algorithm is Proximal Policy Optimization (PPO)~\cite{ppo}, which has been applied in multiple current LLM systems~\cite{ouyang2022training,llama3}
.
Recently, more powerful reasoning models such as KIMI K1.5~\cite{team2025kimi}, Deepseek V3~\cite{deepseekv3}, and R1~\cite{guo2025deepseek} have made modifications to PPO, giving rise to algorithms like GRPO~\cite{grpo} and REINFORCE++~\cite{hu2025reinforceefficientrlhfalgorithm}.
Although these algorithms have shown good performance, their practical deployment is often complicated due to the time-consuming nature of the online exploration involved. 
In contrast, some offline methods~\cite{dpo,kto,zhangonline} are simpler to deploy.
Direct Preference Optimization (DPO) efficiently trains large models for knowledge alignment using preference rankings instead of reward models. 
DPO optimizes classification loss from preference data, making implementing it simpler than RL from human feedback. 
Some papers~\cite{guo2024directlanguagemodelalignment,pang2024iterativereasoningpreferenceoptimization,rao-etal-2025-apt,rao-etal-2025-SEAPO} collectively advance LLM alignment by shifting from static datasets to iterative self-improvement, demonstrating that dynamic, online feedback loops and repeated preference optimization significantly boost both general instruction following and complex reasoning capabilities.
SPHERE~\cite{singh2025selfevolvedpreferenceoptimizationenhancing}, IDPO~\cite{tu2025enhancingllmreasoningiterative} employs a self-evolving, iterative data augmentation approach for mathematical reasoning, called Online DPO. 
Unlike existing work, we improve the effectiveness through the model's self-judgment of the current data selection, rather than the algorithm.

\subsection{Post-training Data Strategies}
Data plays a crucial role in unlocking the capabilities of models~\cite{rao_data_ai,10800533,han-etal-2025-attributes,yu2026mathagent}. In the early days, the LIMA~\cite{zhou2023lima} found that a small amount of data could activate the relevant capabilities of the model and improve the test results of multiple tasks. Recently, some data selections in the field of mathematics have also demonstrated the importance of data quality and diversity. For instance, selections like S1~\cite{muennighoff2025s1simpletesttimescaling} and LIMO~\cite{ye2025limoreasoning}, which used a small amount of data, managed to stimulate the mathematical reasoning capabilities of the models. 
KIMI K1.5~\cite{team2025kimi} adopted curriculum learning and constructed a curriculum-based data training strategy.
Pangu Ultra~\cite{yin2025panguultrapushinglimits} assigned quality and difficulty labels to the data and also used a curriculum-based sampling strategy throughout its three pre-training stages.
In this work, we explored an approach to dynamic data training during the training process, aiming to enhance the final RL performance by selecting training data that is aligned with the model's own competency.

\begin{figure*}[t]
    \centering
    \includegraphics[width=\linewidth]{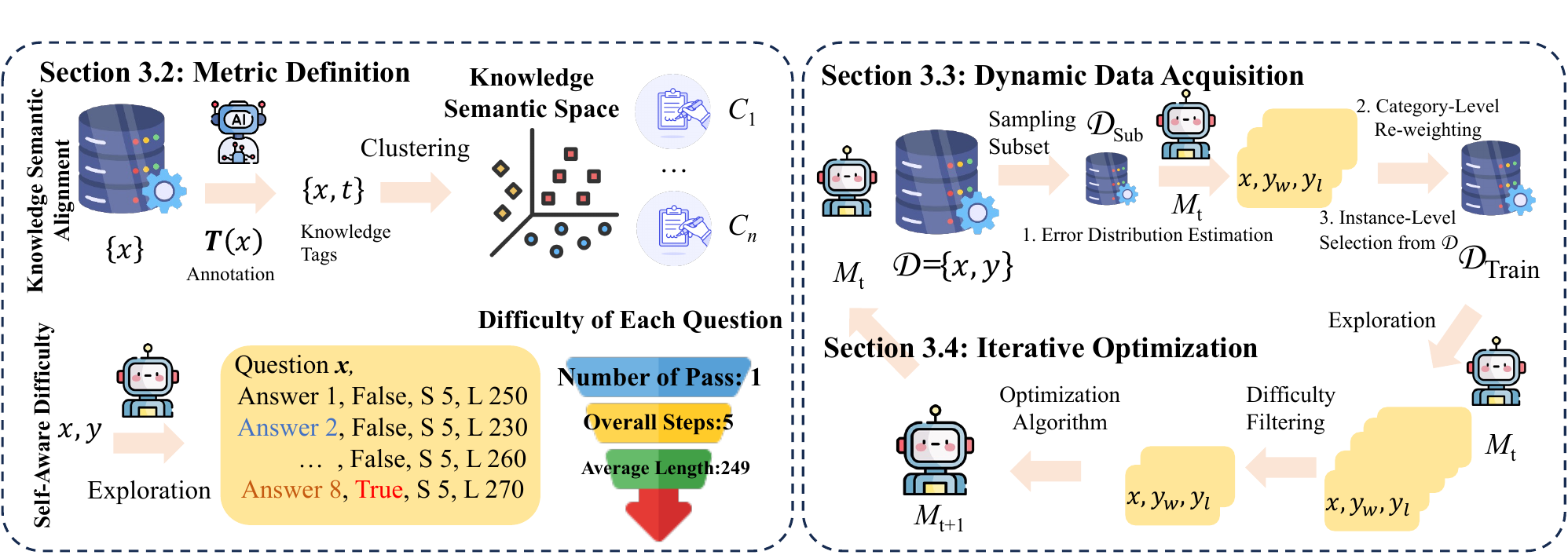}
    \caption{The \method Framework. It operates as an iterative closed-loop consisting of three core phases: (1) Metric Calibration: We first map the dataset into a semantic space using knowledge clustering and simultaneously define instance complexity using the model's self-aware metrics (Pass Rate, Steps, and Length). 
    (2) Dynamic Acquisition:
    Based on a probing subset, we identify weakness domains and re-weight the sampling distribution $P_{adjusted}$ to prioritize high-error clusters and appropriate difficulty levels. (3) Iterative Optimization: The selected curriculum is filtered to remove trivial or intractable samples and used to update the model policy via DPO. The updated model $M_{t+1}$ then re-evaluates the data pool, dynamically shifting the difficulty frontier for the next iteration.    }
    \label{overview}
\end{figure*}
\section{Methods}\label{sec:method}
\subsection{Overview}                       In Figure \ref{overview}, the system operates in cycles. 
At iteration $t$, the current model $M_t$ acts as a probe to evaluate the training data pool $\mathcal{D}$. 
By leveraging two distinct metrics: Knowledge Semantic Alignment and Self-Aware Difficulty, we construct a dynamic curriculum $\mathcal{D}_{train}^{(t)}$ that targets the model's specific weaknesses. 
The model is then updated via preference optimization to yield $M_{t+1}$, shifting the difficulty frontier for the subsequent cycle.
\subsection{Metric Definition}
To operationalize dynamic data selection, we introduce two complementary metrics: one for semantic coverage (what the problem is about) and one for intrinsic complexity (how hard it is for the current model), shown in Figure \ref{overview} (Phase 1).
\subsubsection{Knowledge Semantic Alignment}
Effective training requires diversity across mathematical concepts. We treat knowledge identification as a Latent Semantic Clustering problem.
Annotation: We employ an expert model to generate explicit knowledge tags $T(x)$ for each instance $x$ (e.g., Geometry, Sequence Summation).
Embedding and Clustering: These tags are mapped into a vector space using Sentence-Transformers (all-MiniLM-L6-v2)
\footnote{\url{https://huggingface.co/sentence-transformers/all-MiniLM-L6-v2/tree/main/}}. We then apply K-Means clustering to partition the dataset into $n$ semantic domains $C = \{C_1, C_2, \dots, C_n\}$. This granular partitioning allows us to detect and up-sample specific domains where the model exhibits high error rates. We present some examples in Appendix \ref{sec:diversity_example}.

\subsubsection{Self-Aware Difficulty Calibration}
Unlike external difficulty scorers which are static, we define difficulty as a function of the model's interaction with the data. We propose a \textbf{Hierarchical} Difficulty Metric composed of three dimensions:
    \paragraph{1). Probabilistic Solvability (NoP, Primary):} We perform $K$ explorations for each query. The Number of Passes (NoP), defined as the count of correct responses, serves as the primary proxy for difficulty. A lower NoP indicates higher aleatoric uncertainty and difficulty.  We define the ``solvable range'' as problems where the model is neither consistently correct nor consistently incorrect. 
Specifically, this includes instances with a NoP where $0 < NoP < K$.
    \paragraph{2). Reasoning Depth (Steps, Secondary):} For instances with identical NoP, we differentiate complexity by the number of reasoning steps. We posit that among problems with equal solvability, those necessitating longer logical chains (higher step counts) represent a higher tier of intrinsic complexity. We present examples in Appendix \ref{sec:steps}.
    \paragraph{3). Generation Length:} As a tertiary measure, total token length is used to resolve ties, reflecting the cognitive load of generation.

\subsection{Dynamic Data Acquisition}
Our goal is to shift the training distribution $P(x)$ towards regions of high model uncertainty. This is achieved through a three-step process: Error Distribution Estimation, Category-Level Re-weighting, and Instance-Level Selection.
\paragraph{1). Error Distribution Estimation:}
Since evaluating the entire dataset $\mathcal{D}$ at every iteration is computationally prohibitive, we employ a subset approximation strategy.
Subset Construction: We first sample a small subset $\mathcal{D}_{sub} \subset \mathcal{D}$ that preserves the original cluster proportions. 
This subset acts as a practice exam to probe the current model capability.
Error Identification: The model $M_t$ performs $K$ explorations on $\mathcal{D}_{sub}$. We filter out invalid samples (those consistently correct or consistently incorrect) and identify the Error Dataset $E$, defined as the top 50\% most difficult instances within the solvable range. The distribution of $E$ effectively highlights the semantic clusters where the model currently lacks competency.

\paragraph{2). Category-Level Re-weighting:}
We use the error statistics from $\mathcal{D}_{sub}$ to adjust the sampling probability for the full dataset. Let the dataset be partitioned into $n$ clusters $C_1, \dots, C_n$, with total size $N=\sum |C_i|$. \\
\textbf{Baseline Distribution}: The initial sampling probability $P_{initial}(i)$ for category $i$ reflects the natural data distribution:
\begin{equation}
P_{initial}(i) = \frac{|C_i|}{N}.
\end{equation}
\textbf{Error-Aware Adjustment}: We calculate an adjusted weight $W(i)$ by amplifying categories that appear frequently in the Error Dataset $E$:
\begin{equation}\label{wi}
W(i) = P_{initial}(i) \times (|C_i \cap E| + 1),
\end{equation}

where the term $(+1)$ ensures non-zero smoothing. The final sampling probability for category $i$ becomes $P_{adjusted}(i) = {W(i)}/{\sum W(k)}$. This step systematically biases the sampling distribution towards high-error semantic domains. Equation \ref{wi} focuses on absolute error counts. This is an intentional design to ensure the model maintains performance on the most frequent topics in the distribution.

\paragraph{3). Instance-Level Selection:}
Finally, we convert category-level probabilities into instance-level selection criteria. 
Equation \ref{wij} mitigates bias by normalizing the final instance-level weight $w_{ij}$ by the cluster size $|C_i|$, ensuring that individual hard problems in small clusters still receive significant attention.
For each specific data point $x_{ij}$ belonging to cluster $C_i$, its individual sampling weight $w_{ij}$ is defined as:
\begin{equation}\label{wij}
w_{ij} = \frac{P_{adjusted}(i)}{|C_i|}.
\end{equation}
To construct the final training set $\mathcal{D}_{train}$, we sort all available instances based on their weights $w_{ij}$ and, secondarily, by their intrinsic difficulty scores. We select the top-$S$ instances to form the training set for the current iteration. This mechanism ensures that the model focuses its limited training budget on difficult problems located within its weakest semantic domains.

\subsection{Iterative Optimization}
\paragraph{Difficulty Filtering}
We use rule-based rewards to annotate the model's outputs to obtain correct and incorrect responses. For each problem, we randomly select one correct answer generated by the model as the positive sample and one incorrect answer as the negative sample, thereby constructing a triplet (question $x$, positive answer $y_w$, negative answer $y_l$) to form the training data.
After obtaining the triplet, we can also filter the annotated training samples using the previously defined \textit{Self-aware Difficulty Measurer} to derive the final training data. 
Typically, we first filter out data that is either all incorrect (too difficult) or all correct (too easy), then select the relatively more challenging problems (the top 70\% of the filtered data sorted by difficulty) as the final training set. 

\paragraph{Optimization Algorithm}

Here, we have obtained triplets of training data (question $x$, positive answer $y_w$, negative sample $y_l$) through the previous steps.
We can apply different optimization algorithms, such as the commonly used DPO~\cite{dpo}, or directly use the SFT loss to achieve the RFT (Rejection Sampling Fine-tuning)~\cite{llama3}.
As the model's capabilities improve, ideally, the proportion of difficult training data will gradually decrease.
Eventually, the trainable data will gradually decrease, so that the model's performance reaches a stable value. 
At this point, the iterative process has reached an end.
\subsection{Efficiency Discussion}
Our training efficiency is higher (better results with less data), utilizing only a small amount of additional computation: 1) obtaining the distribution of the subset and 2) measuring the similarity of the overall data. 
For the subset, we only select 1\% data in each iteration for exploratory annotation. 
We dynamically adjust the selection of the training set based on difficulty distribution. 
This part's resource consumption is negligible compared to the total 20,000 training data samples, accounting for less than 1\% of additional time. 
Regarding the measurement of the entire sample dataset, the model labeling we adopt only requires a single annotation process and can be reused. 
The additional time does not exceed 5\% of the overall training process. Generation of 2,000 items takes approximately 10 minutes on our devices. 
Due to the adoption of DPO, the time consumption of our training method is also substantially lower than that of online RL.

\begin{table*}[t!]
  \centering
  \setlength{\tabcolsep}{1mm}
  \scalebox{0.8}{
    \begin{tabular}{lllllllllll}
    \toprule
    \multirow{2}{*}{\textbf{Model}} & \multirow{2}{*}{\textbf{Method}}  
    & \multicolumn{2}{c}{\textbf{Elementary}} & \multicolumn{4}{c}{\textbf{Middle}} &  \multicolumn{2}{c}{\textbf{Competition}}&\multirow{2}{*}{\textbf{Avg.}}\\
    \cmidrule(lr){3-4} \cmidrule(lr){5-8} 
\cmidrule(lr){9-10}&
    &\textbf{GSM} & \textbf{MATH} & \textbf{Minerva} &\textbf{Gaokao} &\textbf{Olympiad} & \textbf{College} &\textbf{Aime24}& \textbf{Amc23}       
  \\
    \midrule
        \multirow{3}{*}{
Llama3.1-8B-Instruct} &\multicolumn{1}{l}{None} & 69.1&	29.0&	16.5&	31.9&	7.3&	21.3&	0.0&	15.3&	23.8
 \\
    & IDPO &76.9\small{$\pm$0.6}&	35.8\small{$\pm$0.3}&19.4\small{$\pm$2.0}&	34.5\small{$\pm$1.0}&	8.5\small{$\pm$0.9}&	23.8\small{$\pm$0.5}&	1.4\small{$\pm$1.7}	&15.1\small{$\pm$0.2}&26.9\small{$\pm$0.4}
	
 \\
    & \quad{+\method} & \textbf{77.4\small{$\pm$0.9}}&	\textbf{36.3\small{$\pm$0.2}}	&\textbf{19.6\small{$\pm$0.6}}&	\textbf{35.7\small{$\pm$0.1}}	&\textbf{10.7\small{$\pm$0.5}}&	\textbf{25.0\small{$\pm$0.1}}&	\textbf{4.0\small{$\pm$0.7}}	&\textbf{20.6\small{$\pm$1.6}}&\textbf{28.7\small{$\pm$0.1}}

 \\

        \midrule

    \multirow{3}{*}{Qwen2.5-7B-Math-Base} & None & 66.7&	64.0&	12.1&	56.1&	28.3&	41.3&	13.9&	41.1&	40.4

 \\    
    & IDPO & 88.4\small{$\pm$0.3}	&72.2\small{$\pm$0.5}&	30.4\small{$\pm$1.7}&	61.5\small{$\pm$1.6}&	38.0\small{$\pm$1.4}&	45.9\small{$\pm$0.2}&	19.4\small{$\pm$2.0}	&60.4\small{$\pm$1.7}&52.0\small{$\pm$0.3}
\\
    & \quad{+\method} & \textbf{89.1\small{$\pm$0.2}}&	\textbf{74.0\small{$\pm$0.2}}&	\textbf{31.7\small{$\pm$1.2}}&	\textbf{62.3\small{$\pm$0.9}}&	\textbf{38.9\small{$\pm$1.7}}&	\textbf{46.3\small{$\pm$0.5}}&	\textbf{23.7\small{$\pm$1.5}}&	\textbf{63.3\small{$\pm$1.2}}
&\textbf{53.7\small{$\pm$0.3}}

 \\
             \midrule
    \multirow{3}{*}{Qwen2.5-7B-Math-SFT} & {None} & 90.8&	73.8&	32.0&	65.5&	37.9&	47.4&	11.7&	65.6	&53.1

 \\
   
    & IDPO &91.3\small{$\pm$0.2}	&\textbf{81.3\small{$\pm$0.3}}&29.9\small{$\pm$0.2}	  &	65.7\small{$\pm$0.9}&	44.7\small{$\pm$0.3}&	\textbf{47.9\small{$\pm$0.1}}&	17.9\small{$\pm$1.1}&	62.9\small{$\pm$0.2}&55.2\small{$\pm$0.1}

    \\
    & \quad+\method & \textbf{91.7\small{$\pm$0.0}}&	81.1\small{$\pm$0.2}&	\textbf{31.3\small{$\pm$0.2}}&	\textbf{67.5\small{$\pm$0.3}}&	\textbf{45.1\small{$\pm$0.9}}&	47.3\small{$\pm$0.2}&	\textbf{19.2\small{$\pm$1.1}}&	\textbf{71.7\small{$\pm$1.5}}&\textbf{56.9\small{$\pm$0.2}}
\\
 \midrule
    \multirow{3}{*}{Qwen3-8B} & {None} & 92.1&	79.0&	34.9&	68.1&	41.3&	44.9&	13.8&	57.2	&53.9

 \\
   
    & IDPO &\textbf{93.0\small{$\pm$0.3}}	&83.4\small{$\pm$2.1}&\textbf{38.8\small{$\pm$0.6}}	  &	71.0\small{$\pm$0.8}&	44.8\small{$\pm$1.3}&	46.1\small{$\pm$0.3}&	23.9\small{$\pm$1.3}&	66.4\small{$\pm$1.2}&58.4\small{$\pm$0.4}

    \\
    & \quad+\method & 92.9\small{$\pm$0.3}&	\textbf{83.9\small{$\pm$0.5}}&	37.6\small{$\pm$0.5}&	\textbf{72.7\small{$\pm$0.9}}&	\textbf{46.0\small{$\pm$1.9}}&	\textbf{46.1\small{$\pm$0.3}}&	\textbf{24.9\small{$\pm$1.6}}&	\textbf{69.2\small{$\pm$0.8}}&\textbf{59.2\small{$\pm$0.1}}\\
    \bottomrule
    \end{tabular}}
    \caption{Main results on multiple test sets for multiple models. 
    The results show that ours can deliver superior results compared to the existing method over multiple model series, notably on competition-level benchmarks.}
  \label{tab:main_res}
\end{table*}
\section{Experiments}\label{main_experiment}
\subsection{Setup}
\paragraph{Baseline and Models}
Following previous work~\cite{guo2024directlanguagemodelalignment,dpo,qwen} of RL, we use Numina-Math~\cite{li2024numinamath}, including 400,000 prompts, as the dataset pool for preference learning dataset construction. 
For fair comparison in the same environment, dataset, and parameters, we compare IDPO~\cite{tu2025enhancingllmreasoningiterative} (iterative training 8 times with random sampling of 20,000 samples per iteration) and PPO (400,000)~\cite{ppo,zeng2025simplerl} as baselines.  
\method is also conducted for 8 iterations in our main experiments.
We use the Qwen2.5 series (7B-Math-Base,7B-Math-SFT
with 15K QwQ~\cite{qwq-32b-preview} data), Qwen3-8B~\cite{yang2025qwen3} and Llama3.1-8B-Instruct as the base models for RL for the main results. We report the average results and standard deviation of three random seeds.
For other analytical experiments, we primarily report the results related to 7B-Math-Base.

\paragraph{Training Details}
The model was trained using a learning rate of $5 \times 10^{-7}$, following a cosine decay strategy for the learning rate schedule. 
A per-device batch size of 1 was used during training, and to achieve an effective total batch size of 128, we employed gradient accumulation over 16 steps, which optimized memory usage during training.
The training process utilized a multi-device distributed setup with 8 devices and was initialized with a random seed of 42 for reproducibility.
To construct the final training set, We select the top-$70\%$ instances to form the training set for the current iteration.
For optimization, we used the Adam optimizer~\cite{kingma2014adam} with $\beta_1 = 0.9$, $\beta_2 = 0.999$, and $\epsilon = 1 \times 10^{-8}$.
Training proceeded for a total of 2 epochs, determined by the total number of training samples and the batch size.
Regarding the hyperparameters of DPO, we set $\beta=0.1$.
For the rollout step in RL, we set the temperature to 1 and perform exploration 8 times, generating 8 responses for each question to obtain positive and negative samples for DPO. 
This value was chosen to provide a stable statistical estimate of the Pass Rate while maintaining manageable inference costs.
We allow a maximum generated length of 3000.

\begin{table}[t!]
    \centering
    \scalebox{0.8}{
    \begin{tabular}{llrc}
        \toprule
        \textbf{Method}&\textbf{Type} & \textbf{Size}& \textbf{Avg.} \\
        \midrule
        LIMO\small{~\cite{ye2025limoreasoning}}&  SFT &0.8K& 46.8\\
        S1\small{~\cite{muennighoff2025s1simpletesttimescaling}}& SFT &1K& 45.9\\
        IRFT\small{~\cite{yuan2023scalingrelationshiplearningmathematical}}&SFT&67K&47.1\\
        PPO\small{~\cite{ppo}}&Online RL&400K &54.6  \\
        IDPO\small{~\cite{zhangonline}}&Offline RL&67K& 52.0 \\
        Ours&Offline RL&{48K}& 53.7 \\
        \bottomrule
    \end{tabular}
    }
    \caption{Compare the training sample utilization efficiency. Qwen2.5-Math-7B serves as the base model. }\label{ppo}
\end{table}

\paragraph{Evaluations}
We follow ~\citet{yang2024qwen25mathtechnicalreportmathematical} using two common English
math benchmarks GSM8K~\cite{gsm}, Math~\cite{hendrycks2021measuringmathematicalproblemsolving}, and using different sets of math tests at different stages to test stronger math skills, such as Minerva Math~\cite{lewkowycz2022solvingquantitativereasoningproblems}, Gaokao
2023 En~\cite{liao-etal-2024-mario}, Olympiad Bench~\cite{he-etal-2024-olympiadbench}, College Math~\cite{tang2024mathscale}, AIME 24
, and AMC 23.
We report greedy
performance on all benchmarks in the zero-shot setting, except for the competition benchmarks (Amc23 and Aime24). 
Considering the limited size of
Amc23 and Aime24, we sample 8 times for each question to mitigate randomness. During the answer generation process for these two datasets, we use a temperature of 0.1 and a top\_p of 0.95.

\subsection{Main Results}
\paragraph{Performance across Different Models}
Table \ref{tab:main_res} establishes the universality of \method across diverse model architectures (Llama-3.1, Qwen-2.5/3) and training paradigms (Base vs. SFT).
Our method consistently outperforms the robust IDPO baseline across all evaluated settings, demonstrating that dynamic data alignment is a fundamental optimizer for reasoning.

Crucially, the performance gains are highly correlated with task difficulty. 
While improvements on simpler datasets (e.g., GSM8K) are marginal due to performance saturation, SAI-DPO unlocks substantial gains on competition-level benchmarks. 
For instance, on the challenging AIME24, our method boosts Qwen2.5-7B-Base from 19.6 (IDPO) to 23.7, and Llama-3.1-8B from 1.4 to 4.0. This trend validates that our error-driven curriculum effectively directs the model's focus toward the frontier of capability, enabling it to master complex reasoning patterns that static baselines fail to capture.
\begin{figure}[t!]
    \centering
    \includegraphics[width=\linewidth]{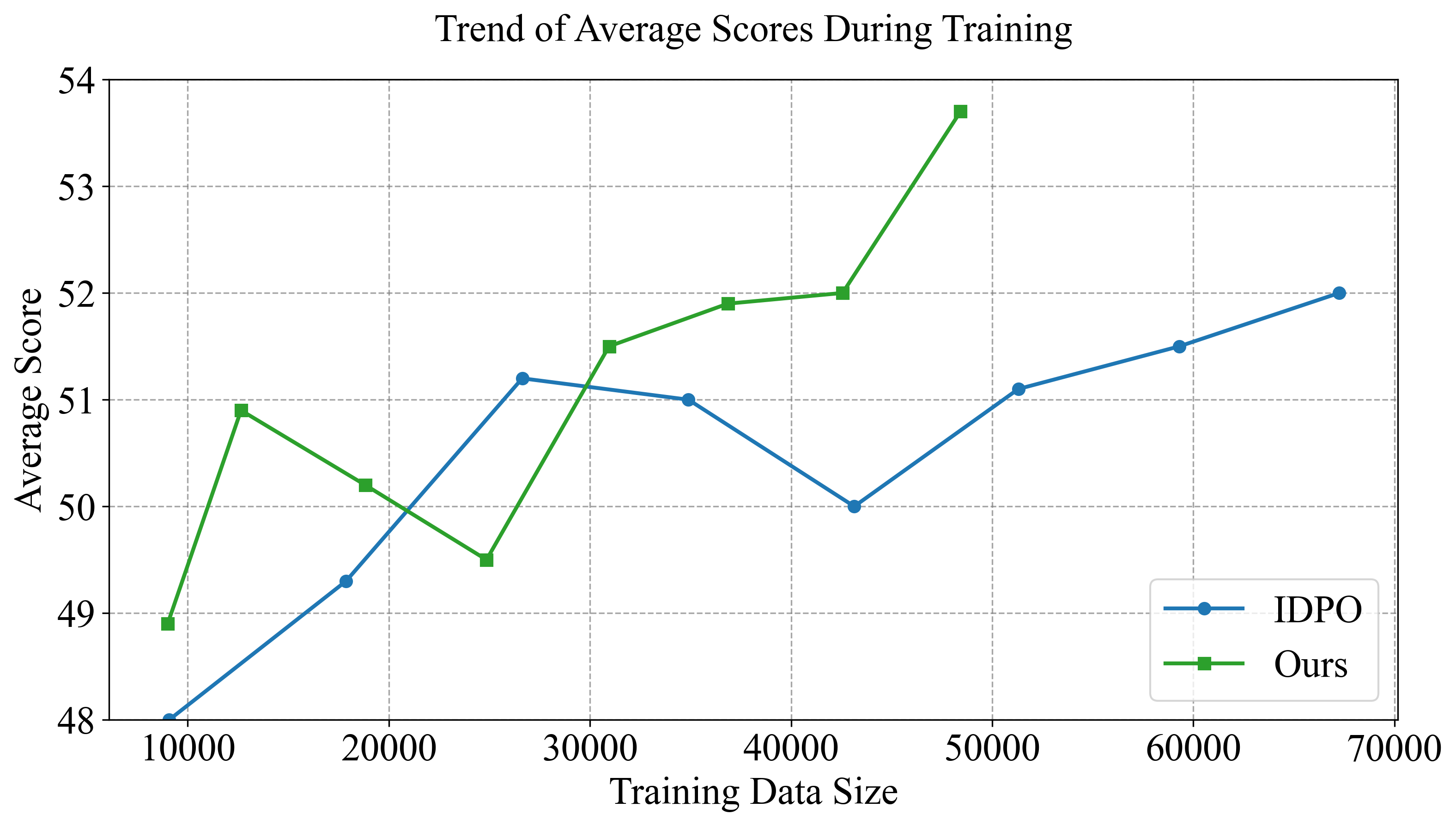}
    \caption{Compare the results of the data scale. Our method achieved better results with fewer data.}
    \label{score_trend}
\end{figure}
\paragraph{Training Sample Utilization Efficiency}
To validate data efficiency, we benchmark \method against three paradigms: Data Distillation (LIMO, S1), Iterative SFT (IRFT), and Online RL (PPO\footnote{ 
\url{https://huggingface.co/RLHFlow/Qwen2.5-7B-PPO-Zero}}). As shown in Table \ref{ppo}, while distillation methods achieve decent baselines with minimal data (1K samples), they suffer from a performance ceiling due to the lack of exploration. 
Conversely, PPO achieves high performance (54.6) but is notoriously data-inefficient, requiring 400K on-policy samples.


\paragraph{Accelerate Convergence}
As shown in Figure \ref{score_trend}, we show the results of the score at the data scale.
Since the amount of training data obtained in each iteration is different, their horizontal coordinates are not the same.
As the amount of training data gradually increases, the overall scores for both methods rise. 
However, the results of our method increased more rapidly and reached a peak earlier. 
This indicates that even with less training data but with difficulty levels matching the model's capabilities, we achieved better results in fewer iterations, highlighting the importance of appropriately difficult samples in training.
These results demonstrate that our method has a higher training efficiency.

\subsection{Ablation Study}\label{sec:ablation}
\paragraph{Dynamic Data Acquisition}
We dismantle the contribution of each component in Table \ref{tab:ablation}. 
The full \method framework achieves the highest average score of 53.7, confirming the synergy of its design.
Impact of Semantic Alignment: Removing the ``Knowledge Semantic Alignment'' module causes the most significant performance degradation ($\delta = -2.3$). This suggests that targeted remedial training on specific weakness domains is the primary driver of our gains.
Hierarchy of Difficulty Features: Within the difficulty modeling, Pass Rate emerges as the dominant factor ($\delta = -1.6$). While fine-grained metrics like Step Count ($\delta = -1.0$) and Length ($\delta = -0.6$) contribute marginally, they act as essential tie-breakers for constructing a smooth curriculum. This validates our hierarchical ranking strategy: statistical uncertainty (Pass Rate) defines the coarse difficulty, while cognitive load (Steps/Length) refines it.
 \begin{table}[t!]
\centering
\scalebox{0.7}
{
\begin{tabular}{llcc}
        \toprule
        \textbf{Component} & \textbf{Variant} & \textbf{Avg.} & \textbf{$\delta$} \\
        \midrule
        \textbf{Full Model} & -- & \textbf{53.7} & -- \\
        \midrule
        Knowledge Semantic Alignment & w/o Similarity & 51.4 & $-2.3$ \\
        \midrule
        \multirow{4}{*}{Self-Aware Difficulty} 
            & w/o Difficulty & 52.5 & $-1.2$ \\
            & w/o Pass Rate  & 52.1 & $-1.6$ \\
            & w/o Step       & 52.7 & $-1.0$ \\
            & w/o Length     & 53.1 & $-0.6$ \\
        \bottomrule
    \end{tabular}
}
\caption{
Ablation study on the defined measurer.}\label{tab:ablation}
\end{table}

\paragraph{Sampling Strategies}
\begin{table}[t!]
\centering
\scalebox{0.8}
{
\begin{tabular}{lcccc}
\toprule
													
    { \multirow{2}{*}{\textbf{Task}}}& \multicolumn{3}{c}{\textbf{Sampling Strategies}}\\
    \cmidrule(lr){2-5} &\textbf{Hard}&\textbf{Curriculum }&\textbf{Clustering}&\textbf{Ours}
    \\\midrule
       GSM &82.8& 84.4&88.5&\textbf{89.1}\\
    MATH &\textbf{74.0}&76.5&71.8&\textbf{74.0} \\
    Minerva  &19.1&18.8&31.2&\textbf{31.7}  \\
    Gaokao &59.5 &60.5&\textbf{62.3}&\textbf{62.3}\\
    Olympiad  &37.5&36.6& \textbf{39.0}&38.9\\
    College&42.2 &42.9&45.8&\textbf{46.3}\\
    Aime24 &\textbf{31.7}&24.6&23.3&23.7\\
    Amc23 &53.8&56.6&60.3&\textbf{63.3}\\
    Avg. &50.1&50.1&52.5&\textbf{53.7}\\
\bottomrule
							
\end{tabular}
}\caption{
Dynamic sampling strategies comparison.
}\label{sampling}
\end{table}
We further compared the following three sampling configurations on Qwen2.5-Math-7B-Base:
1) Hard-mining without tagging (persistent training on the same instances with the most difficult) named ``Hard''~\cite{ye2025limoreasoning},
2) Curriculum learning (progressing sampling from simple to complex) named ``Curriculum''~\cite{chen2025self},
3)  Clustering-only sampling named ``Clustering'' alongside our approach.  
 We only change the sampling strategies. The total amount of data remains the same.
Table \ref{sampling} demonstrates that none of these alternatives outperformed our sampling method, which combines difficulty and similarity.
\paragraph{Difficulty Filtering}

\begin{table}[t!]
\centering
\scalebox{0.8}
{
\begin{tabular}{lcccc}
\toprule

    { \multirow{2}{*}{\textbf{Task}}}& \multicolumn{3}{c}{\textbf{Filtering Strategies}}\\
    \cmidrule(lr){2-5} &\textbf{Easy}&\textbf{Hard}&\textbf{Ours}&\textbf{All}\\\midrule
    Train Samples&34K&34K&48K&67K
    \\\midrule
       GSM &78.4& 86.7&\textbf{89.1}&87.3\\
    MATH &68.4&72.2&\textbf{74.0}&73.0 \\
    Minerva  &21.0&29.4&\textbf{31.7}&29.8  \\
    Gaokao &54.5 &60.8&\textbf{62.3}&60.3\\
    Olympiad  &25.5&37.0& \textbf{38.9}&37.6\\
    College&30.4 &44.9&\textbf{46.3}&44.7\\
    Aime24 &13.7&27.9&23.7&\textbf{34.3}\\
    Amc23 &54.4&58.1&\textbf{63.3}&63.1\\
    Avg. &43.3&52.1&53.7&\textbf{53.8}\\
\bottomrule

\end{tabular}
}\caption{
Compare the results of difficulty filtering. ``Train Samples'' refers to the total number of samples.
}\label{filter}
\end{table}

Table \ref{filter} demonstrates the training results of the data that have been processed using different filtering strategies and then used for training. 
Hard refers to the top 50\% of the most difficult data after directly ranking the acquired data.
Our refers to the top 70\% of the most difficult data after removing the results that are all correct or all wrong and then ranking the remaining data.
All refers to all the data after removing the results that are all correct or all wrong. 
It shows that training with overly difficult data or an excessive amount of easy data can lead to a decline.
A proper mix of data difficulty (including both hard and simple examples) enhances final performance. 
The comparison between our and all demonstrates that using 70\% of the data achieves performance comparable to that of the original data. The simplest portion of the data contributes little to performance.

\section{Analysis}\label{sec:analysis}
\paragraph{Application to Other Post-training Method}
Rejection Sampling Fine-tuning (RFT) is a popular and simple baseline for performing preference fine-tuning, which is performed by many instruction-tuned models~\cite{llama3,qwen}. 
Since this step typically only enhances the base model's capabilities and not those after instruction fine-tuning, we did further comparative experiments on the Qwen-Math-Base model.
From Table \ref{rft}, we can see that combining our data strategy can further improve the effect of RFT with less data. This further demonstrates the generalizability.
\begin{table}[t!]
\centering
\scalebox{0.8}
{
\begin{tabular}{lccccc}
\toprule
    {\textbf{Task}} &\textbf{Base}&\textbf{RFT}&\textbf{RFT+Ours}&$\delta$\\\midrule
    GSM & 66.7&85.7 &\textbf{87.9}&+2.2\\
    MATH &64.0&68.4&\textbf{71.2}&+2.8  \\
    Minerva &12.1&26.1&\textbf{32.4}&+6.3  \\
    Gaokao &56.1&59.0 &\textbf{59.7}&+0.7\\
    Olympiad &28.3&34.1&\textbf{35.6}&+1.5 \\
    College & 41.3&41.5&\textbf{43.2}&+1.7 \\
    Aime24 &13.9&13.8&\textbf{18.4}&+4.6 \\
    Amc23& 41.1&48.1&\textbf{52.2}&+4.1 \\\midrule
    Avg. & 40.4&47.1&\textbf{50.1}&+3.0\\
\bottomrule
\end{tabular}
}
\caption{
Orthogonality among post-training algorithms.}\label{rft}
\end{table}

\paragraph{The Misalignment of External Difficulty}
A critical finding of this work is the orthogonality between external difficulty and model-intrinsic difficulty. In Table \ref{Difficulty}, we compare our method against ``External'' sampling (using a teacher model's score, DeepSeek-R1-Distill-Qwen-14B)
\footnote{\url{https://huggingface.co/deepseek-ai/DeepSeek-R1-Distill-Qwen-14B}}  and P@K-S~\cite{zengb,chen2025self}. Surprisingly, using external difficulty metrics often leads to performance regression (49.9 vs. 52.0 baseline). This implies that a problem deemed hard by a human or a teacher model might be trivial (or hallucinatory) for the student model due to different knowledge boundaries. 
Only Internal (Self-Aware) metrics consistently drive positive transfer, proving that effective curriculum learning must be personalized to the learner's specific latent state.

\begin{table}[t!]
\centering
\scalebox{0.8}
{
\begin{tabular}{lcccc}
\toprule
    {\textbf{Task}} &\textbf{Original}&\textbf{External}&\textbf{P@K-S}&  \textbf{Internal}\\\midrule
    GSM & 88.4&84.4 &87.9&\textbf{89.1}\\
    MATH &72.2&76.4&70.4 &\textbf{74.0 }\\
    Minerva &30.4&18.8&29.4 &\textbf{31.7}  \\
    Gaokao &61.5&60.5 &60.0&\textbf{62.3}\\
    Olympiad &38.0&36.6&38.4 & \textbf{38.9}\\
    College & 45.9&42.9&45.5 &\textbf{46.3}\\
    Aime24 &19.4&\textbf{25.6}&17.9 &23.7\\
    Amc23&60.4&53.8&\textbf{63.4}&63.3\\\midrule
    Avg. & 52.0&49.9&51.6&\textbf{53.7}\\
\bottomrule
\end{tabular}
}
\caption{
Compare the results of difficulty measurement criteria.
External refers to acquiring data from easy to difficult using external difficulty metrics (scoring model) with the human definition.
P@K-S is a comparative metric proposed by ~\citet{zengb}. Internal denotes combining all metrics we defined.}\label{Difficulty}
\end{table}

\paragraph{Numbers of Clusters}
Table \ref{Diversity} investigates how the number of knowledge clusters ($N$) affects performance. The model achieves the best results with $N=150$ (Avg. 53.7), while deviating from this number leads to degradation: Too Few Clusters ($N=50$): Performance drops to 52.1. With overly broad categories, distinct knowledge points are merged, preventing the system from accurately targeting specific weaknesses.
Too Many Clusters ($N=200, 250$): Increasing granularity beyond the optimal point hurts performance. The score decreases to 53.3 at $N=200$ and drops further at $N=250$. 
Excessive fragmentation results in sparse data per category, causing unstable difficulty estimation and reducing the effectiveness of the sampling strategy

\begin{table}[t!]
\centering
\scalebox{0.8}
{
\begin{tabular}{lrrrr}
\toprule
    { \multirow{2}{*}{\textbf{Task}}}& \multicolumn{4}{c}{\textbf{Number of Categories}}\\
    \cmidrule(lr){2-5} &\textbf{50}&\textbf{150}&\textbf{200}& \textbf{250}\\\midrule
       GSM & 87.1&\textbf{89.1}&88.6&88.2\\
    MATH &70.4&\textbf{74.0}&72.8&70.0 \\
    Minerva  &27.6&\textbf{31.7}&26.5&30.9  \\
    Gaokao  &\textbf{62.6}&62.3&61.6&60.5\\
    Olympiad  &\textbf{39.1}& 38.9&\textbf{39.1}&37.9\\
    College &43.9&\textbf{46.3}&45.8&46.0\\
    Aime24 &26.7&23.7&26.7&\textbf{27.1}\\
    Amc23 &59.4&63.3&\textbf{65.0}&58.1\\\midrule
    Avg. &52.1&\textbf{53.7}&53.3&52.3\\
\bottomrule
\end{tabular}
}\caption{
Compare the impacts of the number of clusters. 
}\label{Diversity}
\end{table}
\paragraph{Dependence on Knowledge Tags Annotation Model}
We further investigate the sensitivity of \method to the quality of semantic tags by scaling the expert annotator  (DeepSeek-R1-Distill-Qwen from 7B to 32B) (Table \ref{expert_scale}). It demonstrates a robust positive correlation between annotator capability and downstream performance, with the average score rising from 52.4 to 54.4. 
Notably, this benefit is disproportionately concentrated on complex reasoning benchmarks; for instance, performance on AIME24 surges from 21.2 to 30.0. 
This suggests that highly capable annotators are essential for accurately disentangling deep semantic structures in challenging problems, thereby ensuring the curriculum precisely targets the model's cognitive blind spots. 
We employ the 14B expert in other experiments with minimal additional compute.

\begin{table}[t!]
\centering
\scalebox{0.8}
{
\begin{tabular}{lrrr}
\toprule
    { \multirow{2}{*}{\textbf{Task}}}& \multicolumn{3}{c}{\textbf{Expert Model Scale}}\\
    \cmidrule(lr){2-4} &\textbf{7B}&\textbf{14B}& \textbf{32B}\\\midrule
       GSM & 88.6&\textbf{89.1}&88.2\\
    MATH &73.0&\textbf{74.0}&73.4 \\
    Minerva  &31.2&\textbf{31.7}&31.6  \\
    Gaokao  &62.1&\textbf{62.3}&61.2\\
    Olympiad  &\textbf{39.1}& 38.9&37.9\\
    College &45.9&\textbf{46.3}&45.6\\
    Aime24 &21.2&23.7&\textbf{30.0}\\
    Amc23 &62.8&63.3&\textbf{67.2}\\\midrule
    Avg. &52.4&53.7&\textbf{54.4}\\
\bottomrule
\end{tabular}
}\caption{
Compare the impacts of the Expert Model Scale. Superior model capabilities yield better results.
}\label{expert_scale}
\end{table}

\begin{figure}[t]
    \centering
    \includegraphics[width=\linewidth]{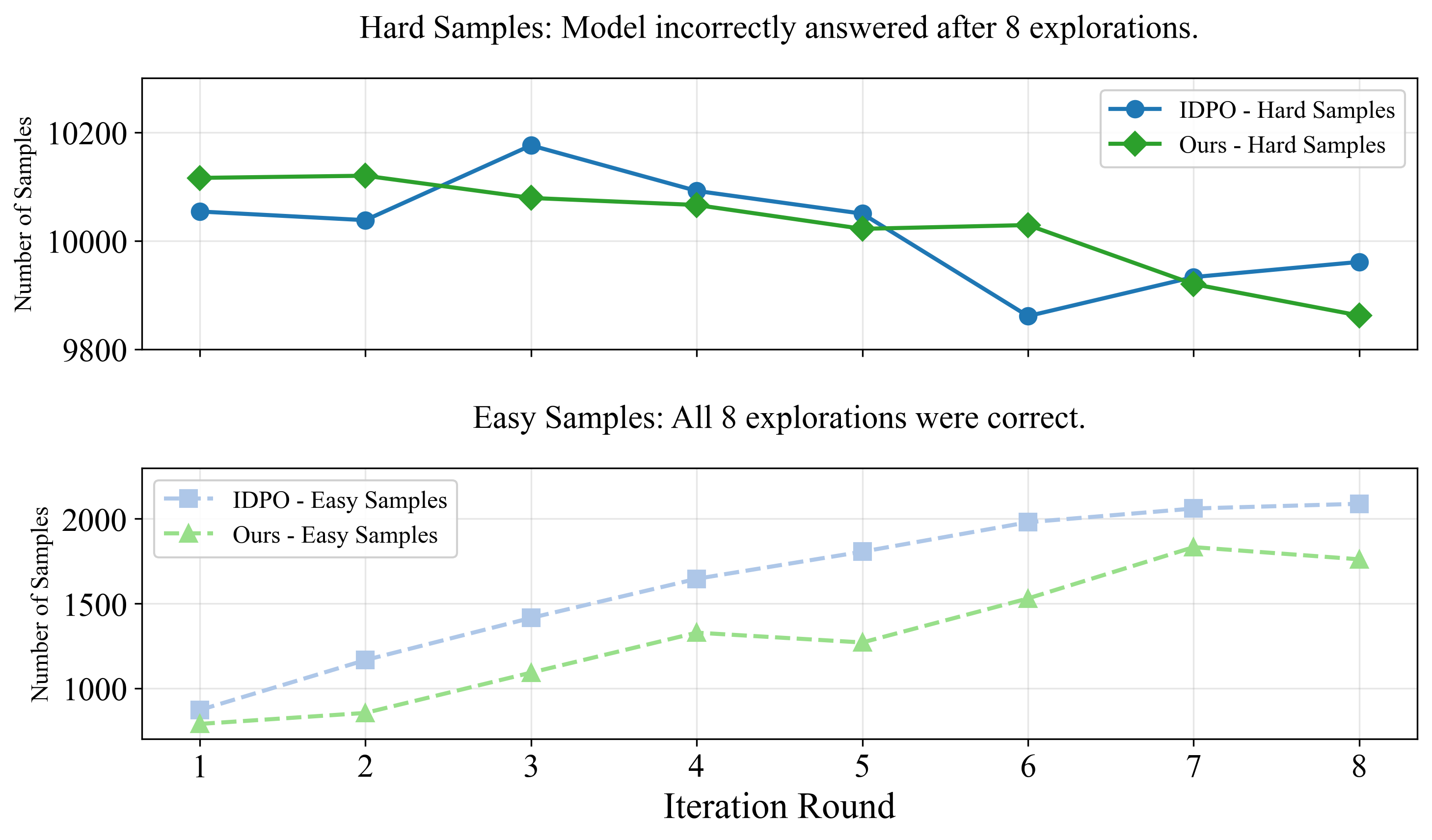}
    \caption{Comparison of the ability to solve sample variability. Both our method and IDPO reduce unsolvable hard samples and enhance simple samples. }\label{hard_samples}
\end{figure}

\paragraph{Benefits of Hard Samples}
How does the curriculum evolve? Figure \ref{hard_samples} visualizes the dynamic composition of the training set over iterations. 
Initially, the sampler identifies a large volume of Hard samples (high uncertainty). As training progresses, the model digests these instances, converting them from Hard (low NoP) to Easy (high NoP), causing the volume of remaining Hard samples to decrease. 
Crucially, \method maintains a higher solving rate for hard problems compared to IDPO. This demonstrates our method's superior ability to learn effectively from sampled data.

\section{Conclusion}
This work challenges the prevailing paradigm of static data selection in mathematical reasoning. By introducing \method, we demonstrate that the definition of high-quality data is inherently dynamic and model-dependent. Our framework, which couples semantic clustering with self-aware difficulty metrics, enables models to autonomously navigate their learning trajectory. 
The significant gains observed on AIME24 and AMC23 suggest that future research should pivot from simply scaling dataset size to optimizing the alignment between data complexity and real-time model capability.

\section*{Limitations}
There are several limitations to our work.
First, due to training resource constraints, our algorithm primarily focuses on offline RL methods such as iterative DPO and does not conduct sampling-related experiments on mainstream online approaches like PPO.  
Another limitation is that because of the capacity ceiling of offline algorithms, our final performance does not surpass mainstream online RL algorithms like PPO.  
Finally, we have only conducted experiments on test sets of varying difficulty levels within the field of mathematical reasoning, and other domains such as code and general domains remain unexplored. 
%
\section*{Ethics Statement}
Our work follows the ACL Ethics Policy. Our findings are based on publicly available datasets for reproducibility purposes. 
All procedures performed in this study are in accordance
with the ethical standards. 
In addition, it is hard to control the generation of LLMs. 
We should be aware of the potential problems caused by hallucinations.

\section*{Acknowledgments}
This work was supported in part by Guangdong S\&T Program (Grant No. 2024B0101050003), Guangdong Basic and Applied Basic Research Foundation (Grant No. 2024A1515011491), and Shenzhen Science and Technology Program (Grant Nos. ZDSYS20230626091203008, KJZD20231023094700001, KQTD20240729102154066). 
We would like to thank the anonymous reviewers and meta-reviewer for their insightful suggestions.

\bibliography{custom}

\appendix
\newcounter{checksubsection}
\newcounter{checkitem}[checksubsection]

\newcommand{\checksubsection}[1]{%
  \refstepcounter{checksubsection}%
  \paragraph{\arabic{checksubsection}. #1}%
  \setcounter{checkitem}{0}%
}

\newcommand{\checkitem}{%
  \refstepcounter{checkitem}%
  \item[\arabic{checksubsection}.\arabic{checkitem}.]%
}
\newcommand{\question}[2]{\normalcolor\checkitem #1 #2 \color{blue}}
\newcommand{\ifyespoints}[1]{\makebox[0pt][l]{\hspace{-15pt}\normalcolor #1}}

\section{Appendix}
\label{sec:appendix}

\subsection{Training Times Comparison}
The following table summarizes the total wall-clock training time for 8 iterations, comparing our SAI-DPO framework with the IDPO baseline. While SAI-DPO introduces a dynamic sampling phase, its superior data efficiency leads to a significant reduction in total training time.

\begin{table}[htbp]
  \centering
  \caption{Training Efficiency Comparison of Different Methods}
  \label{tab:efficiency}
  \resizebox{\linewidth}{!}{
  \begin{tabular}{llcc}
    \toprule
    Model Scale & Method          & Total Data Volume & Avg. Time per Iteration \\
    \midrule
    \multirow{2}{*}{7B Model} & IDPO (Baseline) & 67K               & 7.0 hours               \\
                              & SAI-DPO (Ours)  & 48K               & $\sim$5.3 hours         \\
    \midrule
    \multirow{2}{*}{8B Model} & IDPO (Baseline) & 67K               & 8.0 hours               \\
                              & SAI-DPO (Ours)  & 48K               & $\sim$6.0 hours         \\
    \bottomrule
  \end{tabular}
  }
\end{table}
\subsection{Examples of Tags}\label{sec:diversity_example}
We present several examples for different types of math problems. 
We use a uniform prompt template to prompt the model to output the corresponding tag. The specific prompt template is as follows: ``<|begin of sentence|><|User|>What knowledge points need to be involved in solving the following questions. Answer should be output in the following format, no need to output the answer, reply in English, \#\#\#Knowledge Points:\{\}$\textbackslash$nPlease output the results directly, reducing the thought process.\{input\}<|Assistant|>\#\#\#Knowledge Points:'' 
The \{input\} is a placeholder indicating the input question.
As shown in the figure (Tag Case), the second and third examples both involve trigonometry, indicating repetitive knowledge points. By leveraging this tagging, we can better identify data with similar knowledge points, thereby enabling self-learning by locating example problems for knowledge points where the model currently has weaknesses.

\subsection{Examples of Problem-Solving Steps}\label{sec:steps}

We present the output results of several models, which can automatically continue to generate replies in this format by adding the field ``Step: 1''.
We present response examples from the Qwen-7B-Distill model. As shown in the examples, the model can output logical problem-solving steps and include reflections (e.g., ``wait''), where correct reflection processes solve problems using fewer logical steps. Additionally, the output results can be used to calculate total step counts, response length, and correctness, thereby determining the difficulty of the current data for the model.
\onecolumn
\renewcommand{\thefigure}{A.\arabic{figure}}
\setcounter{figure}{0}
\begin{tcolorbox}[colback=cyan!5!white, colframe=cyan!75!black, title=
Responses of Explorations]
    \textbf{Question:} 

    A girl has three different types of coins in her wallet: $3$ nickels, $3$ dimes, and $2$ quarters. She also adds $2$ half-dollars. She randomly picks two coins, calculates the total value, and puts them back. She repeats this process multiple times. What is the maximum number of different sums she can record?

    \textbf{gt:10}

    \textbf{False Response 1 (No answer output):}
    
     \textcolor{blue}{Step 1}: Identify all possible coin combinations.
     First, list all the coins available:- 3 nickels (each worth $0.05$)- 3 dimes (each worth $0.10$)- 2 quarters (each worth $0.25$)- 2 half-dollars (each worth $0.50$)...
    
    \textcolor{blue}{Step 2}: Determine all possible sums for picking two coins. ...
    
    \textcolor{red}{Wait}, perhaps not directly.

\textcolor{red}{Wait}, she picks two coins, calculates their sum, puts them back, and repeats.

So, each pick results in a sum, which is recorded.

She does this multiple times, perhaps until she gets all possible sums.

\textcolor{red}{Wait}, but the problem seems to suggest that the sums are recorded based on how many times they occur, so perhaps it's about the frequency.

But, in any case, we need to find out how

    \textbf{False Response 2 (Wrong answer):}
    
\textcolor{blue}{Step 1}: Identify the possible values of each type of coin ...

\textcolor{blue}{Step 2}: Determine all possible combinations of two coins being picked.
- Nickels: There are 3 nickels, but since the girl is picking two coins at random, she could pick two nickels, a nickel and a dime, a nickel and a quarter, a nickel and a half-dollar, a dime and a quarter, a dime and a half-dollar, and a quarter and a half-dollar.

\textcolor{blue}{Step 3}: Calculate the unique sums for each combination ...

\textcolor{blue}{Step 4}: Ensure that all sums are unique.
Looking at the sums: $0.10$, $0.15$, $0.30$, $0.55$, $0.35$, $0.60$, $0.75$. These sums are all distinct.

\textcolor{blue}{Step 5}: Determine the number of different sums.
From the above, we have:
- $0.10$ from two nickels
- $0.15$ from nickel and dime
- $0.30$ from nickel and quarter
- $0.55$ from nickel and half-dollar
- $0.35$ from dime and quarter
- $0.60$ from dime and half-dollar
- $0.75$ from quarter and half-dollar

So, there are 7 different sums.

Therefore, the maximum number of different sums she can record is $\boxed{7}$ ...

**Final Answer**: 
$\boxed{7}$

    \textbf{True Response 1:}

    \textcolor{blue}{Step 1}: Determine the total number of coins and their values ...

So, total coins: 3 + 3 + 2 + 2 = 10 coins.

\textcolor{blue}{Step 2}: Calculate all possible sums of two coins.

We need to consider all combinations of picking two coins and summing their values, then find out how many unique sums are possible.

\textcolor{blue}{Step 3}: List all possible pairs and their sums ...

So, it seems like these are all the possible sums.

Therefore, the number of different sums is 10.

\textcolor{red}{Wait}, but is there any overlap or additional sums I might be missing?

Let me think differently.

...

Hence, the maximum number of different sums is 10.

Therefore, the final answer is: $\boxed{10}$

I hope it's correct.

\#\# **Final Answer**
$\boxed{10}$
I think that's the answer.

\end{tcolorbox}

\clearpage

\begin{tcolorbox}[colback=pink!5!white, colframe=pink!75!black, title=Tag Case 1]
    



    \textbf{Question:} 
    Evaluate the expression $5^{\log_{5}(9+4)}$.

    \textbf{Tags:} 
    Logarithmic and Exponential Expressions and Equations

\end{tcolorbox}

\begin{tcolorbox}[colback=pink!5!white, colframe=pink!75!black, title=Tag Case 2]
    \textbf{Question:} 
    Determine the domain of the function $f(x) =\tan(\arcsin(x^2)).$
    
    \textbf{Tags:}
    Trigonometric functions, inverse trigonometric functions, domain and range

\end{tcolorbox}

\begin{tcolorbox}[colback=pink!5!white, colframe=pink!75!black, title=Tag Case 3]
\textbf{Question:}

Given the function $f(x) = A \sin(\omega x + \phi)$, where $(A > 0, \omega > 0, |\phi| < \frac{\pi}{2})$, its graph intersects the $y$-axis at $(0, \frac{3}{2})$, and its first highest and lowest points on the right side of the $y$-axis are $(x\_0, 3)$ and $(x\_0 + 2\pi, -3)$ respectively.

1. Find the analytical expression of the function $y = f(x)$.

2. How can the graph of this function be obtained through translation and scaling transformations from the graph of $y = \sin x (x \in \mathbb{R})$?

3. Find the intervals where this function is monotonically increasing and its center of symmetry.
\textbf{Tags:}
Trigonometric functions, function transformation, monotonicity, symmetry, maximum and minimum points.
\end{tcolorbox}
\begin{tcolorbox}[colback=pink!5!white, colframe=pink!75!black, title=Tag Case 4]
    \textbf{Question:} 
Find the largest prime divisor of the number $102111011_6$.

    \textbf{Tags:}
     Number Bases, Prime Factorization.

\end{tcolorbox}

\end{document}